\documentclass{article}
\usepackage{ijcai26}
\usepackage{amsmath,bm}
\usepackage[table]{xcolor}
\usepackage{amsfonts}

\definecolor{darkgreen}{rgb}{0.0, 0.7, 0.0}

\definecolor{darkgray}{HTML}{F69337}
\colorlet{medgray}{darkgray!65}  
\colorlet{lightgray}{darkgray!30}

\def\eqref#1{equation~\ref{#1}}

\def\1{\bm{1}}

\def\vd{{\bm{d}}}

\def\vo{{\bm{o}}}
\def\vp{{\bm{p}}}

\DeclareMathAlphabet{\mathsfit}{\encodingdefault}{\sfdefault}{m}{sl}
\SetMathAlphabet{\mathsfit}{bold}{\encodingdefault}{\sfdefault}{bx}{n}

\usepackage{times}
\usepackage{soul}
\usepackage{url}
\usepackage[hidelinks]{hyperref}
\usepackage[utf8]{inputenc}
\usepackage[small]{caption}
\usepackage{graphicx}
\usepackage{amsmath}
\usepackage{amsthm}
\usepackage{booktabs}
\usepackage{algorithm}
\usepackage{algorithmic}
\usepackage[switch]{lineno}
\usepackage{multirow}
\usepackage{tikz}
\usepackage{makecell}
\usepackage[toc,page]{appendix}

\title{Visual Implicit Geometry Transformer for Autonomous Driving}

\author{
Arsenii Shirokov$^1$\and
Mikhail Kuznetsov$^1$\and
Danila Stepochkin$^1$\and
Egor Evdokimov$^1$\and
Daniil Glazkov$^1$\and
Nikolay Patakin$^1$\and
Anton Konushin$^1$\and
Dmitry Senushkin$^1$\\
\affiliations
$^1$Lomonosov Moscow State University\\
\emails
}

\begin{document}

\maketitle

\begin{abstract}
    We introduce the Visual Implicit Geometry Transformer (ViGT), an autonomous driving geometric model that estimates continuous 3D occupancy fields from surround-view camera rigs. ViGT represents a step towards foundational geometric models for autonomous driving, prioritizing scalability, architectural simplicity, and generalization across diverse sensor configurations. Our approach achieves this through a calibration-free architecture, enabling a single model to adapt to different sensor setups. 
    Unlike general-purpose geometric foundational models that focus on pixel-aligned predictions, ViGT estimates a continuous 3D occupancy field in a bird’s-eye-view (BEV) addressing domain-specific requirements.
    ViGT naturally infers geometry from multiple camera views into a single metric coordinate frame, providing a common representation for multiple geometric tasks.
    Unlike most existing occupancy models, we adopt a self-supervised training procedure that leverages synchronized image-LiDAR pairs, eliminating the need for costly manual annotations.
    We validate the scalability and generalizability of our approach by training our model on a mixture of five large-scale autonomous driving datasets (NuScenes, Waymo, NuPlan, ONCE, and Argoverse) and achieving state-of-the-art performance on the pointmap estimation task, with the best average rank across all evaluated baselines. 
    We further evaluate ViGT on the Occ3D-nuScenes benchmark, where ViGT achieves comparable performance with supervised methods. The source code is publicly available at \href{https://github.com/whesense/ViGT}{https://github.com/whesense/ViGT}.   
\end{abstract}

\begin{figure}[t]
    \centering
    \resizebox{\linewidth}{!}{
\definecolor{colorVGGT}{HTML}{3B82F6}
\definecolor{colorDUSt3R}{HTML}{C21A09}
\definecolor{colorDA3}{HTML}{1F2937}
\definecolor{colorStream3R}{HTML}{F97316}
\definecolor{colorOurs}{HTML}{22C55E}
\definecolor{gridcolor}{HTML}{E5E7EB}
\definecolor{labelcolor}{HTML}{374151}
\definecolor{tickcolor}{HTML}{9CA3AF}
\definecolor{legendgray}{HTML}{6B7280}
\definecolor{ourstext}{HTML}{166534}

\def\graphlinewidth{2.0}
\def\legendlinewidth{1.5}
\def\chartradius{160}

\def\legendtext{\normalsize}
\def\ticktext{\small}
\def\labeltext{\Large}

\begin{tikzpicture}[scale=0.03]

\foreach \tickval in {1,2,3,4,5,6,7} {
    \pgfmathsetmacro{\r}{\tickval / 7 * \chartradius}
    \draw[gridcolor, line width=0.5pt] (0,0) circle (\r);
    \node[tickcolor, font=\ticktext\selectfont, anchor=south west] at (4    , \r) {\tickval};
}


\foreach \i/\dataset in {0/NuScenes, 1/Waymo, 2/ONCE, 3/NuPlan, 4/AV2} {
    \pgfmathsetmacro{\angle}{90 - \i * 72}
    \draw[gridcolor, line width=0.5pt] (0,0) -- (\angle:\chartradius);
    \node[labelcolor, font=\labeltext\selectfont\bfseries, anchor=center] at (\angle:\chartradius + 20) {\dataset};
}





\pgfmathsetmacro{\rNuScenes}{5.61 / 7 * \chartradius}
\pgfmathsetmacro{\rWaymo}{5.43 / 7 * \chartradius}
\pgfmathsetmacro{\rONCE}{5.36 / 7 * \chartradius}
\pgfmathsetmacro{\rNuPlan}{4.30 / 7 * \chartradius}
\pgfmathsetmacro{\rAV}{4.49 / 7 * \chartradius}
\draw[colorDA3, line width=\graphlinewidth pt, fill=none] 
    (90:\rNuScenes) -- (18:\rWaymo) -- (-54:\rONCE) -- (-126:\rNuPlan) -- (-198:\rAV) -- cycle;

\pgfmathsetmacro{\rNuScenes}{4.47 / 7 * \chartradius}
\pgfmathsetmacro{\rWaymo}{3.53 / 7 * \chartradius}
\pgfmathsetmacro{\rONCE}{6.12 / 7 * \chartradius}
\pgfmathsetmacro{\rNuPlan}{4.59 / 7 * \chartradius}
\pgfmathsetmacro{\rAV}{3.91 / 7 * \chartradius}
\draw[colorDUSt3R, line width=\graphlinewidth pt, fill=none] 
    (90:\rNuScenes) -- (18:\rWaymo) -- (-54:\rONCE) -- (-126:\rNuPlan) -- (-198:\rAV) -- cycle;

\pgfmathsetmacro{\rNuScenes}{3.93 / 7 * \chartradius}
\pgfmathsetmacro{\rWaymo}{2.63 / 7 * \chartradius}
\pgfmathsetmacro{\rONCE}{7.19 / 7 * \chartradius}
\pgfmathsetmacro{\rNuPlan}{4.33 / 7 * \chartradius}
\pgfmathsetmacro{\rAV}{3.99 / 7 * \chartradius}
\draw[colorStream3R, line width=\graphlinewidth pt, fill=none] 
    (90:\rNuScenes) -- (18:\rWaymo) -- (-54:\rONCE) -- (-126:\rNuPlan) -- (-198:\rAV) -- cycle;

\pgfmathsetmacro{\rNuScenes}{4.02 / 7 * \chartradius}
\pgfmathsetmacro{\rWaymo}{2.38 / 7 * \chartradius}
\pgfmathsetmacro{\rONCE}{6.23 / 7 * \chartradius}
\pgfmathsetmacro{\rNuPlan}{3.70 / 7 * \chartradius}
\pgfmathsetmacro{\rAV}{3.72 / 7 * \chartradius}
\draw[colorVGGT, line width=\graphlinewidth pt, fill=none] 
    (90:\rNuScenes) -- (18:\rWaymo) -- (-54:\rONCE) -- (-126:\rNuPlan) -- (-198:\rAV) -- cycle;

\pgfmathsetmacro{\rNuScenes}{1.81 / 7 * \chartradius}
\pgfmathsetmacro{\rWaymo}{2.43 / 7 * \chartradius}
\pgfmathsetmacro{\rONCE}{5.82 / 7 * \chartradius}
\pgfmathsetmacro{\rNuPlan}{3.30 / 7 * \chartradius}
\pgfmathsetmacro{\rAV}{2.97 / 7 * \chartradius}
\draw[colorOurs, line width=\graphlinewidth pt, fill=colorOurs, fill opacity=0.25] 
    (90:\rNuScenes) -- (18:\rWaymo) -- (-54:\rONCE) -- (-126:\rNuPlan) -- (-198:\rAV) -- cycle;


\node[anchor=north] at (0, - \chartradius - 5) {
    \begin{tikzpicture}[scale=1]
        \draw[colorVGGT, line width=\legendlinewidth pt] (0, 0) rectangle (0.4, 0.25);
        \node[legendgray, font=\legendtext, anchor=west] at (0.5, 0.125) {VGGT};
        
        \draw[colorDUSt3R, line width=\legendlinewidth pt] (1.8, 0) rectangle (2.2, 0.25);
        \node[legendgray, font=\legendtext, anchor=west] at (2.3, 0.125) {DUSt3R};
        
        \draw[colorDA3, line width=\legendlinewidth pt] (3.8, 0) rectangle (4.2, 0.25);
        \node[legendgray, font=\legendtext, anchor=west] at (4.3, 0.125) {DA3};
        
        \draw[colorStream3R, line width=\legendlinewidth pt] (5.4, 0) rectangle (5.8, 0.25);
        \node[legendgray, font=\legendtext, anchor=west] at (5.9, 0.125) {Stream3R};
        
        \fill[colorOurs, fill opacity=0.25] (7.6, 0) rectangle (8.0, 0.25);
        \draw[colorOurs, line width=\legendlinewidth pt] (7.6, 0) rectangle (8.0, 0.25);
        \node[legendgray, font=\legendtext\bfseries, anchor=west] at (8.1, 0.125) {Ours};

    \end{tikzpicture}
};

\end{tikzpicture}
}
    \caption{Our Visual Implicit Geometry Transformer outperforms the most recent fundamental geometric models on publicly available autonomous driving datasets in Chamfer Distance$\downarrow$. Positions closer to the center indicate better performance.}
\end{figure}
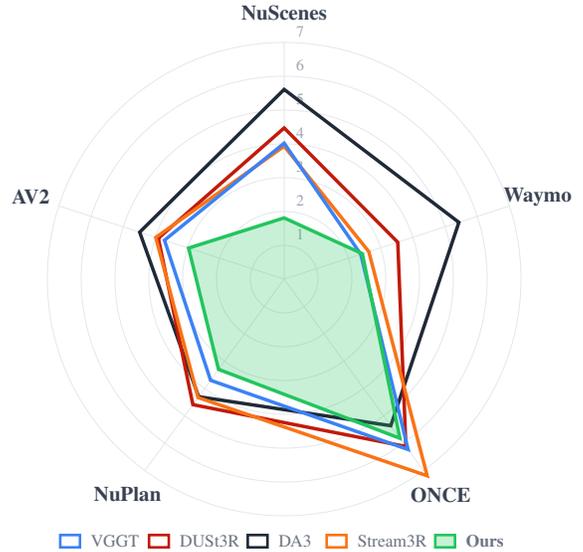

\section{Introduction}
\label{sec:introduction}

Accurate, metric geometric perception models are crucial for reliable and safe autonomous driving. Learning 3D-aware geometric representations from images and other sensor modalities has become central to progress in both autonomous driving and general scene understanding.

Recent advances in multi-view reconstruction~\cite{dust3r2023,depthanythingv32025} have enabled a new class of end-to-end geometric models that directly estimate dense 3D point maps from uncalibrated image collections, implicitly recovering camera parameters. While these methods achieve strong performance on monocular and multi-view reconstruction in general domain, their application to autonomous driving scenarios remains challenging. Firstly, surround-view camera rigs commonly used in autonomous vehicles prioritize full $360^\circ$ coverage with limited redundancy, resulting in reduced overlap between camera views and weakening the multi-view geometric assumptions underlying these approaches. Secondly, many recent multi-view reconstruction models do not explicitly enforce accurate metric scale, which is critical for autonomous driving applications. Thirdly, per-pixel depth and point predictions are typically aligned with individual images rather than the global scene, making it difficult to reconcile inconsistencies across views. In contrast, discretized 3D voxel representations that explicitly encode occupied and free space have become widely adopted in modern autonomous driving systems due to their direct compatibility with downstream tasks such as collision checking and motion planning. However, voxel-based approaches often face scalability challenges, as they typically require dense and costly ground-truth annotations for training.

Occupancy fields provide an attractive alternative by modeling 3D scenes as continuous functions over space, rather than relying on explicit voxel grids or per-view depth maps. Such representations are compact, differentiable, resolution- and sensor-agnostic, and naturally align geometry from multiple views into a unified scene-centric coordinate frame, making them well suited for autonomous driving applications.

In this work, we introduce the Visual Implicit Geometry Transformer, which estimates a continuous 3D occupancy field directly from surround-view camera images. Our model learns a compact continuous bird's-eye-view~(BEV) field that can be decoded into binary occupancy using a point-wise decoder. This field provides a flexible geometric representation of the scene and can be further transformed into downstream representations by querying point-wise occupancy probabilities and accumulating them through rendering.
In contrast to previous approaches, we simplify the image-to-BEV transformation by making it calibration-free and fully data-driven. This design reduces inductive bias and improves the generality and scalability of the model. Finally, we adopt a scalable, sensor-agnostic, self-supervised training procedure that leverages synchronized but uncalibrated multi-view images and LiDAR data without requiring additional annotations.

\section{Related Work}
\label{sec:related_work}

\paragraph{General geometric models.}

Foundational geometric models have been an active area of research for several decades. Early works introduced learning-based monocular depth estimation~\cite{eigen2014} and subsequently demonstrated that scaling such models with large and diverse datasets~\cite{midas2022,Patakin_2022_CVPR} leads to improved generalization. In particular, pixel-aligned depth prediction models trained with calibrated~\cite{eigen2014}, uncalibrated~\cite{midas2022,Patakin_2022_CVPR}, or metric supervision~\cite{bhat2023zoedepth} were shown to produce geometry estimates that generalize effectively to in-the-wild scenarios. However, monocular depth estimation remains fundamentally ill-posed due to depth ambiguity, which limits generalization when relying on single-view inputs~\cite{yin2023metric}. To address this limitation, recent works proposed transformer-based architectures~\cite{dust3r2023,mast3r,monst3r} that predict dense point maps instead of per-pixel depth. By leveraging multi-view supervision during training, these models reconstruct image-aligned visible geometry~\cite{dust3r2023,mast3r,depthanythingv32025} and, in some cases, estimate camera parameters from uncalibrated image collections~\cite{vggt}. Subsequent approaches~\cite{cut3r,point3r2025} introduced latent memory embedding that can be rendered from novel viewpoints, albeit with a limited number of views. While these methods significantly improve scalability with respect to both training data and the number of input images, they still rely heavily on multi-view correspondences and therefore exhibit substantial performance degradation under low-overlap image collections~\cite{rig3r2025}. Moreover, most existing approaches regress scale-invariant depth or point representations, which are insufficient for recovering 3D metric volumetric geometry required by downstream applications such as autonomous driving. In this work, we introduce a scalable transformer-based model that estimates a continuous 3D occupancy field at metric scale directly from surround-view images. Differently, our method is designed for surround-view driving scenarios with limited camera overlap, alleviating multi-view constraints.

\begin{figure*}[!t] 
    \centering
    \includegraphics[width=\textwidth]{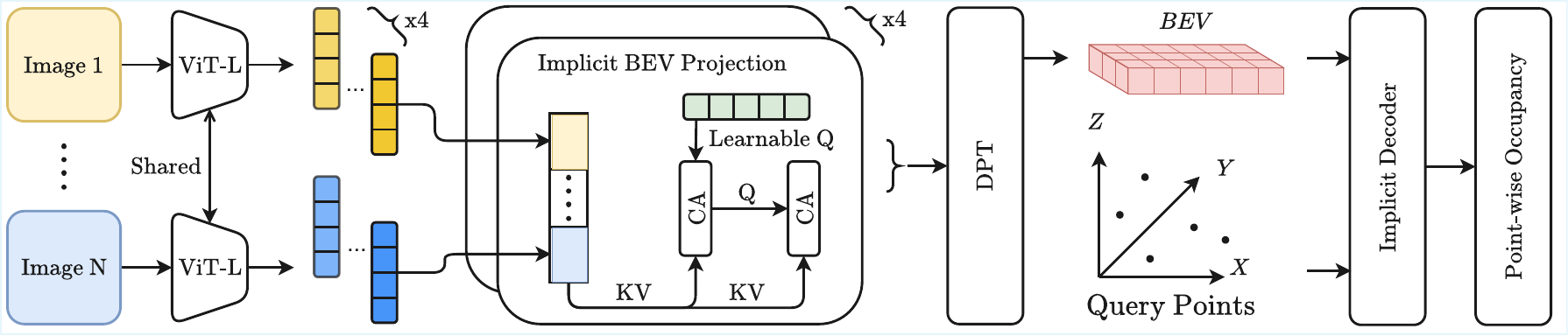}
    \caption{Our architecture consists of three main components: (1) an image encoder (ViT-L) that independently processes each image and extracts feature tokens from the last four layers, producing four sequences of tokens per image; (2) a calibration-free Implicit BEV Projection module that projects tokens from each encoder layer across all images to their corresponding BEV space, generating four layer-specific BEV representations, which are then aggregated and upsampled into a single unified BEV representation using DPT; and (3) a query-based Implicit Decoder that predicts occupancy probabilities for 3D points from the final BEV features. This design enables pure data-driven scene modeling without geometric inductive biases.}
    \label{fig:model}
\end{figure*}
\paragraph{Geometric models for autonomous driving.} 

Prior geometric models for autonomous driving have predominantly focused on semantic occupancy estimation, as it provides a structured representation suitable for downstream tasks. Vision-based perception models in this domain build upon earlier works~\cite{bevformer2022,lss2020} that introduced forward~\cite{bevformer2022} and backward~\cite{lss2020} mappings from images to bird’s-eye-view (BEV) space, where subsequent estimations are performed. More recent methods~\cite{marinello2025offsetocc,pan2024renderocc,huang2024selfocc,sparseocc2024} typically lift multi-view image features into voxel grids using explicit geometric projections~\cite{bevformer2022,lss2020}, followed by refinement through 3D convolutional or transformer-based fusion~\cite{zhou2022cross}. Alternative approaches~\cite{liu2022bevfusion,zhang2024fusionocc} incorporate explicit LiDAR encoders to improve prediction accuracy. Despite their effectiveness, these architectures rely on precise camera calibration, which limits generalization to novel camera configurations or scenarios with calibration errors. Additionally, they require expensive annotations, such as voxel-level semantic labels, which constrains scalability. In contrast, we introduce the simple transformer-based model architecture that does not use camera calibration for BEV construction. Our model learns an continuous 3D occupancy field in BEV representation that incorporates depth, point clouds, and occupancy from images, enabling scalable and transferable 3D perception across diverse self-driving datasets.

Recent advances~\cite{uno2024,ALSO2023,vidar2023} have introduced continuous occupancy forecasting in both space and time, leveraging pretext geometric tasks to support downstream applications. Early methods~\cite{ALSO2023,uno2024,4docc2023} operate on LiDAR point clouds and employ convolutional architectures to predict binary occupancy at the point level. Subsequent works~\cite{uno2024,Liu2023occ4cast} extend this paradigm by incorporating temporal modeling. Similarly, visual point cloud forecasting methods, such as ViDAR~\cite{vidar2023}, predict future LiDAR-like point clouds from monocular image sequences rather than estimating occupancy directly. However, the majority of these models remain biased by calibration-dependent architectures, limiting their ability to generalize across diverse sensor configurations and environments. We leverage continuous occupancy estimation task for training our models since it provides annotation-free supervision from raw data.

\section{Method}

In this work, we introduce the Visual Implicit Geometry Transformer~(ViGT) that estimates continuous 3D occupancy field directly from surround-view camera images. The remainder of this section is organized as follows: section~\ref{subsec:model} presents ViGT architecture that incorporates scalable blocks without inductive biases; section~\ref{subsec:uno2024e} describes unsupervised occupancy estimation problem as the scalable task for training; finally, section~\ref{subsec:loss} details the label-free training procedure enabling learning continuous 3D occupancy fields end-to-end.

\subsection{Visual Implicit Geometry Transformer}
\label{subsec:model}

We step towards general foundational geometric models for autonomous driving by designing our model architecture, prioritizing scalability, simplicity, and generalization across diverse sensor configurations.
Our architecture consists of three main components (Fig.~\ref{fig:model}): an image encoder that independently processes each image, a calibration-free implicit camera-to-BEV projection module that unifies independent image features in a scene-centric BEV representation and a query-based implicit decoder that predicts occupancy values for 3D points.

\paragraph{Image Encoder.}

Following principles of simplicity and scalability, we employ a ViT-Large backbone~\cite{vit} to extract visual features from camera images. In contrast to prior autonomous driving models that rely on convolutional neural networks with limited receptive fields~\cite{bevformer2022,lss2020}, transformer architectures capture long-range contextual dependencies through self-attention while imposing minimal inductive bias. Each image is encoded independently by the ViT-Large model, producing a compact set of token embeddings that serve as input to subsequent processing stages.

\begin{figure*}[t]
    \centering
    \includegraphics[width=\textwidth]{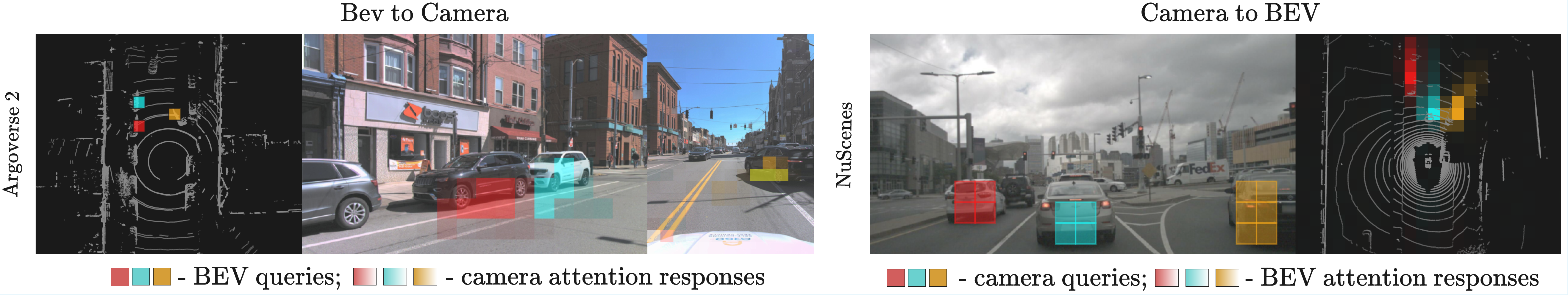}
    \caption{Cross-attention visualization demonstrating consistent correspondences between BEV queries and camera tokens. 
    (Left) Image attention heatmaps for selected BEV queries,
    (Right) BEV attention heatmaps for selected image queries.
    These visualizations confirm that the implicit BEV projection learns geometrically correct camera-to-BEV transformations.}
    \label{fig:implicit_bev}
\end{figure*}

\paragraph{Implicit BEV projection.}

Autonomous driving applications require scene-centric geometric representations defined at metric scale within a unified coordinate frame. Unlike image-aligned representations, such as depth or point maps, scene-centric representations enable direct reasoning about spatial occupancy, which is critical for autonomous driving tasks. To this end, we adopt a bird’s-eye-view (BEV) representation, as it provides comprehensive scene coverage while being more computationally and memory efficient than dense 3D volumetric grids. Existing self-driving approaches~\cite{bevformer2022,lss2020} rely on known camera intrinsics and extrinsics to explicitly project image features into BEV space. Guided by principles of simplicity and scalability, we seek a calibration-free transformation from multi-view visual features to BEV space that can be applied across diverse camera setups.

We design our implicit camera-to-BEV projection module to learn the geometric transformation from data (Figure~\ref{fig:implicit_bev}, Figure~\ref{fig:rig}). Specifically, the implicit camera-to-BEV projection employs two sequential cross-attention blocks between image patches and BEV latent queries. The number of BEV latent queries is smaller than the number of image patch tokens, addressing computationally efficiency. 
To capture multiscale geometric information, we project outputs of the four last layers of ViT-L encoder independently to BEV space, using our implicit BEV projection with non-shared weights and individual BEV queries. The resulting multi-layer BEV representations are then aggregated and upscaled using DPT~\cite{dpt2021} to obtain the final fine-grained scene-level BEV representation. We validate our design choices through ablation studies comparing different projection structures and encoder layer selections (Table~\ref{tab:ablation}).

\begin{figure}[t]
    \centering
    \includegraphics[width=\linewidth]{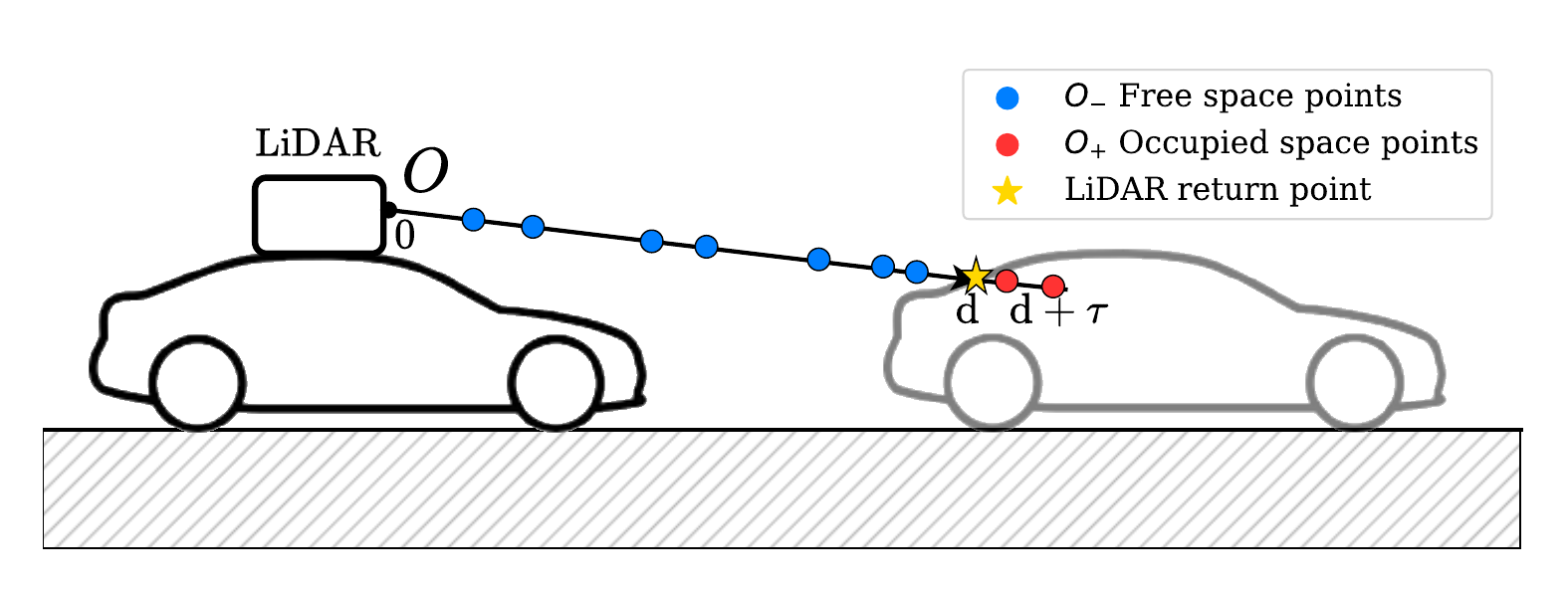}
    \caption{We construct two sets of training points sampled along each LiDAR ray. Points before the reflection point are labeled as free space (negative label), while points near the reflection point correspond to occupied space (positive label).}
    \label{fig:imocc_adv}
\end{figure}

\paragraph{Implicit decoder.}

To transform the discrete BEV grid into a continuous occupancy field, we employ an implicit decoder inspired by ImplicitIO~\cite{implicitio} that maps 3D query points to occupancy probabilities. We linearly interpolate query features and concatenate them with normalized query points. The concatenated features are decoded using Convolutional Occupancy Network \cite{peng2020convolutional} to predict the occupancy probabilities for the query point. 

\subsection{Unsupervised Occupancy Estimation}
\label{subsec:uno2024e}

To build a scalable training framework, we leverage a training objective that relies exclusively on data that can be collected efficiently on real-world autonomous platforms. Specifically, we assume the availability of synchronized multi-view images and corresponding LiDAR point clouds. Unlike conventional occupancy estimation approaches that operate on discretized voxel grids, we avoid explicit spatial discretization and instead formulate the task in continuous space, operating directly on points. LiDAR is a high-precision active sensor that emits laser pulses and measures the return time from reflecting surfaces to estimate distances.
By construction, LiDAR measurements guarantee that the space between the sensor origin and the first surface intersection along a ray is free. Accordingly, we construct two sets of training points sampled along each LiDAR ray. 
Points located before the reflection point are labeled as free space and thus assigned negative occupancy labels, while points near reflection point correspond to solid surfaces and are assigned positive occupancy labels. More formally, given a normalized LiDAR ray parameterized by origin~$\vo$ and direction~$\vd$, we define two occupancy classes as follows:
\begin{equation}
\begin{cases}
    O_{-}: \vp_{i} = \vo + t_{i}\vd, \quad t_{i} \in [0, d], \\
    O_{+}: \vp_{j} = \vo + t_{j}\vd, \quad t_{j} \in [d, d+\tau],
\end{cases}
\end{equation}
where $d$ is the distance from the ray origin $\vo$ to the reflected surface and $\tau$ is the thickness hyperparameter that regulates the minimal surface depth along rays. This value is usually small for models that should perceive thin surfaces.

Overall, the training objective is formulated as a binary occupancy classification problem, where the model predicts the occupancy label of each queried 3D point given the available input context. Prior works~\cite{ALSO2023,uno2024} primarily rely on single or multiple LiDAR point clouds as input, resulting in LiDAR-only models. In contrast, we adopt a vision-only paradigm and leverage this training objective to supervise the Visual Implicit Geometry Transformer.

\subsection{Training}
\label{subsec:loss}
Following the unsupervised occupancy estimation formulation in Section~\ref{subsec:uno2024e}, we sample query points along LiDAR rays to construct positives (occupied space) and negatives (free space) occupancy labels. We employ a balanced sampling strategy, that ensures equal representation of negative  and positive targets. We divide the ray segment $[0, d)$ into $K$ bins and sample uniformly within each bin, ensuring coverage across the entire free space. However, simple uniform sampling for negative queries can lead to inefficient supervision near object surfaces, where accurate boundary predictions are crucial. So, to better utilize object borders and improve surface localization, we augment this with symmetric negative sampling from the region $[d-\tau, d)$ where $\tau$ is the same hyperparameter used for positive sampling. This symmetric approach samples negative points near the surface boundary, providing explicit supervision for the model to learn object borders. We validate our query sampling strategy through ablation studies comparing random, stratified, and stratified with symmetric interval sampling (Table~\ref{tab:ablation}). 

We train the model using binary cross-entropy loss. For a batch of query points $\{\vp_i\}_{i=1}^{N}$ with ground truth occupancy labels $\{y_i \in \{0,1\}\}_{i=1}^{N}$ and predicted occupancy probabilities $\{o_i \in [0,1]\}_{i=1}^{N}$, the loss is defined as:
\begin{equation}
    \mathcal{L} = -\frac{1}{N} \sum_{i=1}^{N} \left[ y_i \log(o_i) + (1-y_i) \log(1-o_i) \right].
\end{equation}

\subsection{Rendering}

Our model estimates an continuous occupancy field, which provides a flexible geometric representation that can be transformed into various downstream representations by querying point-wise occupancy probabilities and accumulating them through volume rendering. This unified representation enables conversion to point clouds and occupancy voxel grids, facilitating application to diverse autonomous driving tasks while maintaining a single, consistent geometric model.

\paragraph{Ray-based rendering.} 
To extract point clouds, we leverage volumetric integration to render LiDAR rays. Given the ray with origin $\vo$ and direction $\vd$, we sample points $\{\vp_i = \vo + t_i \vd\}_{i=1}^{N}$, where $t_i$ are evenly spaced sampling distances. We decode occupancy probability $o(\vp_i)$ for every point and integrate the depth value accordingly: 
\begin{equation}
    \label{eq:ray_rendering}
    d = \sum_{i=1}^{N} t_i \cdot o(\vp_i) \cdot T_i, \quad T_i = \prod_{j=1}^{i-1} (1 - o(\vp_j)),
\end{equation}
where $T_i$ is the transmittance accumulated from predicted occupancy.

\paragraph{Voxel grid rendering.} 
To generate binary voxel grids from continuous occupancy field, we query M points randomly sampled within each voxel and aggregate their occupancy predictions. For a voxel $V$ the occupancy is computed as the maximum response across all predictions:
\begin{equation}
    O(V) = \max_{j=1}^{M} o(\vp_j),
\end{equation}

\begin{figure}[t]
    \includegraphics[width= \linewidth]{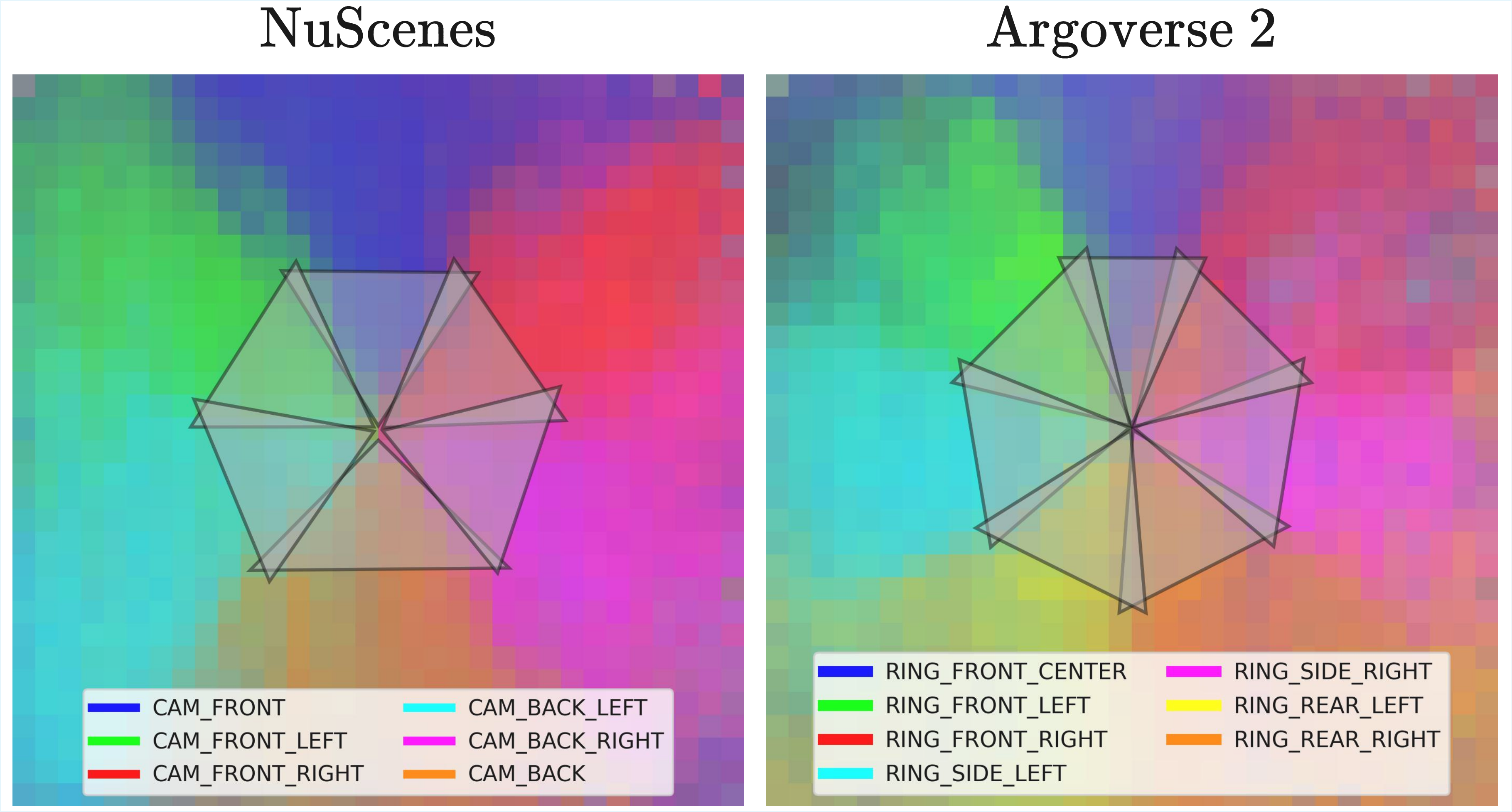}
    \caption{The camera-to-bev attention responses (colored regions) accurately partition BEV space according to each camera's field of view, demonstrating that the implicit BEV projection correctly learns to map image features to their corresponding spatial sectors in BEV space without explicit camera parameters. Shown for different camera rigs configuration.}
    \label{fig:rig}
\end{figure}
\section{Experiments}
\subsection{Training details}
We train the model with the AdamW optimizer for 200K iterations, using a cosine learning rate scheduler with a peak learning rate of $5 \times 10^{-5}$ and a warmup of 10K iterations. To ensure diverse training data, we use balanced sampling across five autonomous driving datasets: NuScenes~\cite{caesar2020nuscenes}, ONCE~\cite{mao2021one}, Waymo~\cite{sun2020scalability}, Argoverse 2~\cite{wilson2023argoverse}, and NuPlan~\cite{caesar2021nuplan}. For each batch, we sample equally from all datasets to prevent bias towards larger datasets. We use a batch size of 6 per GPU. The input images are resized with the shortest side set to 192 pixels. We use 1024 BEV queries with 1024 channels in each Implicit BEV projector, aggregating features to 256×256 resolution with 256 channels after DPT. The training runs on 128 A100 GPUs over five days. We employ gradient norm clipping with a threshold of 1.0 to ensure training stability.

\begin{table}[t]
    \centering
    \resizebox{0.45\textwidth}{!}{%
    \begin{tabular}{llcccc}
        \toprule
        Method & Conf. & Supervision & Calib. & F1-score$\uparrow$ & IoU$\uparrow$ \\
        \midrule
        FB-Occ & ICCV 2023 & 3D GT & + & \cellcolor{medgray}{0.8181} & \cellcolor{medgray}{0.7022} \\
        Sparse-Occ & ECCV 2024 & 3D GT & + & 0.6271 & 0.4680 \\
        PanoOcc & CVPR 2024 & 3D GT & + & \cellcolor{darkgray}{0.8347} & \cellcolor{darkgray}{0.7271} \\
        Offset-Occ & CVPR 2025 & 3D GT & + & 0.6240 & 0.4637 \\
        \midrule
        RenderOcc & ICRA 2024 & 2D GT & + & 0.6442 & 0.4824 \\
        \midrule
        Self-Occ & CVPR 2024 & Self-sup & + & 0.6552 & 0.4960 \\
        \midrule
        Ours & & Self-sup & - & \cellcolor{lightgray}{0.7115} & \cellcolor{lightgray}{0.5658} \\
        \bottomrule
    \end{tabular}%
    }
    \caption{\textbf{Occ3D-NuScenes}. Our method achieves \textbf{third-best overall performance} and \textbf{first-best perfomance} over previous self-supervised methods and methods with 2d labels supervision. Our method is the only approach which don't use camera calibration.}
    \label{tab:results_occ}
\end{table}

\subsection{Implicit BEV projection}
We demonstrate that our calibration-free implicit BEV projection correctly learns geometric transformations without explicit camera parameters. Figure~\ref{fig:implicit_bev} illustrates attention responses between BEV queries and camera tokens, showing consistent correspondences in both forward (camera-to-BEV) and inverse (BEV-to-camera) attention directions. Figure~\ref{fig:rig} demonstrates that the projection correctly aligns image features to their corresponding spatial sectors in BEV space for different camera rigs (6 cameras on NuScenes and 7 cameras on Argoverse 2) from different datasets, with each camera's field of view accurately mapped to its correct frustum region. These figures empirically confirm that the implicit projection successfully learns geometrically correct camera-to-BEV transformations in a calibration-free manner across diverse datasets with different camera rig configurations.

\begin{table*}[t]
    \centering
    \small
    \resizebox{\textwidth}{!}{%
    \begin{tabular}{llcccccccccccc}
        \toprule
        Method & Conf. & Rendering & \multicolumn{2}{c}{NuScenes} & \multicolumn{2}{c}{AV2} & \multicolumn{2}{c}{Waymo} & \multicolumn{2}{c}{ONCE} & \multicolumn{2}{c}{NuPlan} & \makecell{Avg.\\Rank$\downarrow$} \\
        \cmidrule(lr){4-5} \cmidrule(lr){6-7} \cmidrule(lr){8-9} \cmidrule(lr){10-11} \cmidrule(lr){12-13} \cmidrule(lr){14-14}
        & & & AbsRel$\downarrow$ & CD$\downarrow$ & AbsRel$\downarrow$ & CD$\downarrow$ & AbsRel$\downarrow$ & CD$\downarrow$ & AbsRel$\downarrow$ & CD$\downarrow$ & AbsRel$\downarrow$ & CD$\downarrow$ \\
        \midrule

        \multirow{2}{*}{VGGT} & \multirow{2}{*}{CVPR 2025} & D  &
            0.195 & 4.019 &
            \cellcolor{lightgray}{0.164} & \cellcolor{lightgray}{3.724} &
            \cellcolor{medgray}{0.084} & \cellcolor{darkgray}{2.381} &
            0.450 & 6.230 &
            0.167 & 3.696 & \cellcolor{lightgray}{3.9} \\
        & & P & 
            0.303 & 4.794 &
            0.220 & 4.305 &
            0.113 & 3.576 &
            0.568 & 7.566 &
            0.215 & 5.047 \\
        \midrule

        \multirow{2}{*}{DUSt3R} & \multirow{2}{*}{CVPR 2024} & D  &
            0.250 & 4.466 &
            0.230 & 3.906 &
            0.152 & 3.534 &
            0.515 & 6.124 &
            0.236 & 4.590 & 6.9 \\
        & & P & 
            0.375 & 5.321 &
            0.445 & 6.309 &
            0.347 & 7.726 &
            0.887 & 10.341 &
            0.508 & 6.749 \\
        \midrule

        \multirow{2}{*}{Mast3R} & \multirow{2}{*}{ECCV 2024} & D  &
            \cellcolor{lightgray}{0.191} & 4.583 &
            0.190 & 4.752 &
            0.120 & 2.949 &
            0.498 & 8.402 &
            0.181 & 4.826 & 6.4 \\
        & & P & 
            0.285 & 5.487 &
            0.355 & 6.876 &
            0.260 & 4.673 &
            0.843 & 10.906 &
            0.635 & 8.475 \\
        \midrule

        \multirow{2}{*}{Monst3R} & \multirow{2}{*}{ICLR 2025} & D  &
            0.217 & 3.945 &
            0.201 & 4.013 &
            0.127 & 3.037 &
            \cellcolor{lightgray}{0.390} & \cellcolor{darkgray}{5.039} &
            0.182 & \cellcolor{lightgray}{3.443} & 4.8 \\
        & & P & 
            0.415 & 7.024 &
            0.279 & 5.726 &
            0.210 & 5.442 &
            1.308 & 15.659 &
            0.263 & 7.221 \\
        \midrule

        \multirow{2}{*}{Stream3R} & \multirow{2}{*}{---} & D  &
            \cellcolor{medgray}{0.173} & \cellcolor{lightgray}{3.931} &
            \cellcolor{medgray}{0.162} & 3.992 &
            \cellcolor{darkgray}{0.083} & 2.629 &
            0.465 & 7.190 &
            \cellcolor{lightgray}{0.163} & 4.331 & 4 \\
        & & P & 
            0.397 & 5.985 &
            0.237 & 5.799 &
            0.114 & 3.522 &
            0.705 & 10.859 &
            0.207 & 5.668 \\
        \midrule

        \multirow{2}{*}{Cut3R} & \multirow{2}{*}{CVPR 2025} & D  &
            0.189 & 4.062 &
            0.194 & \cellcolor{medgray}{3.681} &
            \cellcolor{lightgray}{0.101} & \cellcolor{lightgray}{2.478} &
            0.447 & 5.970 &
            \cellcolor{medgray}{0.156} & \cellcolor{darkgray}{3.000} & \cellcolor{medgray}{3.5} \\
        & & P & 
            0.240 & 5.233 &
            0.342 & 6.760 &
            0.176 & 3.866 &
            0.811 & 11.442 &
            0.172 & 3.688 \\
        \midrule

        DA3 & --- & D  &
            0.289 & 5.606 &
            0.174 & 4.488 &
            0.383 & 5.426 &
            0.403 & \cellcolor{medgray}{5.360} &
            0.265 & 4.300 & 6.4 \\
        \midrule  

        RenderOcc & ICRA 2024 & PR & 
            0.245 & \cellcolor{medgray}{3.637} &
            0.316 & 10.864 &
            0.526 & 8.442 &
            \cellcolor{medgray}{0.298} & 7.490 &
            0.332 & 5.295 & 7.3\\
        \midrule
        
        {\textbf{Ours}} &  & PR & 
            \cellcolor{darkgray}{0.068} & \cellcolor{darkgray}{1.807} &
            \cellcolor{darkgray}{0.131} & \cellcolor{darkgray}{2.965} &
            0.121 & \cellcolor{medgray}{2.431} &
            \cellcolor{darkgray}{0.169} & \cellcolor{lightgray}{5.821} &
            \cellcolor{darkgray}{0.118} & \cellcolor{medgray}{3.298} & \cellcolor{darkgray}{1.8}\\
        \bottomrule
    \end{tabular}%
    }
    \caption{\textbf{Pointmap Estimation.} Our method achieves \textbf{the best Average Rank (1.8)} across all metrics and datasets, demonstrating the best overall performance. Rendering types: D denotes depth map unprojection from camera views, P denotes direct point cloud from pointmaps, and PR denotes points rendering from predicted occupancy fields.}
    \label{tab:results}
\end{table*} 

\subsection{Occupancy estimation}
The first experiment compares our model with recent geometric models designed for autonomous driving applications.

\paragraph{Evaluation protocol.}
We evaluate all models on the Occ3D-NuScenes benchmark~\cite{tian2023occ3d} which comprises 150 validation scenes with voxelized occupancy annotations defined within a region of interest of $[-40\text{m},-40\text{m},-1\text{m}]\times[+40\text{m}, +40\text{m}, +5.4\text{m}]$ in ego-vehicle coordinate frame with voxel resolution $[0.4\text{m}, 0.4\text{m}, 0.4\text{m}]$. Unlike the original benchmark, which provides semantic occupancy annotations across 17 categories, we focus exclusively on geometric occupancy estimation. To this end, all semantic classes are mapped to a single occupied label, reducing the task to binary occupancy prediction. We evaluate occupancy estimation using two metrics: F1-score and Intersection over Union (IoU). Both metrics are computed over the binary occupancy predictions (occupied vs. free) within the camera's visible region. 

\paragraph{Baselines.}
We compare our method against recent vision-based occupancy prediction approaches as shown in Table~\ref{tab:results_occ}. All models are grouped by their supervision type. Most approaches employ dense voxel grids with explicit 3D occupancy annotations: FB-Occ~\cite{li2023fb}, PanoOcc~\cite{wang2024panoocc}, Sparse-Occ~\cite{liu2023sparseocc}, and Offset-Occ~\cite{marinello2025offsetocc}. Methods with reduced supervision include RenderOcc~\cite{pan2024renderocc}, which trains with 2D supervision, and Self-Occ~\cite{huang2024selfocc}, which learns occupancy in a self-supervised manner without voxel labels, similar to our approach. 

\paragraph{Experimental results.} 
We evaluate our model without voxel finetuning to demonstrate that our learned representation can generalize to new geometric tasks. Our model achieves third-best overall performance with an F1-score of $0.7115$ and IoU of $0.5658$, while being the only approach that is both fully self-supervised and calibration-free. Unlike top-performing methods such as PanoOcc and FB-Occ that require expensive 3D annotations, our method uses only raw LiDAR point clouds for supervision, eliminating the need for costly manual labeling.
Compared to the previous best self-supervised method (Self-Occ), we achieve improvements of $0.056$ in F1-score and $0.07$ in IoU, while additionally removing the calibration requirement.

\subsection{Pointmap Estimation}
Second, we evaluate our model against general-domain vision-based methods that are similarly calibration-free but produce pixel-aligned predictions.
\paragraph{Evaluation protocol.}

We evaluate all methods on the public test splits of widely used autonomous driving datasets, including NuScenes, Argoverse 2, Waymo, ONCE, and NuPlan, using LiDAR point clouds as ground-truth targets. Due to the large scale of nuPlan, we subsample the official test split by selecting every 100th LiDAR frame for evaluation. We report two evaluation metrics: Absolute Relative Error (AbsRel$\downarrow$), computed along LiDAR rays, and Chamfer Distance (CD$\downarrow$). For pixel-aligned methods, predictions are sampled at image pixels corresponding to the projections of LiDAR points onto the image plane.
To provide an aggregate measure of overall performance across all datasets and metrics, we compute the Average Rank (Avg Rank$\downarrow$) for each method as the mean of performance places (1st, 2nd, 3rd, 4th, etc.) assigned to each metric-dataset combination, where lower rank values indicate better overall performance.

\begin{figure*}[!ht]
    \centering
    \includegraphics[width=\linewidth]{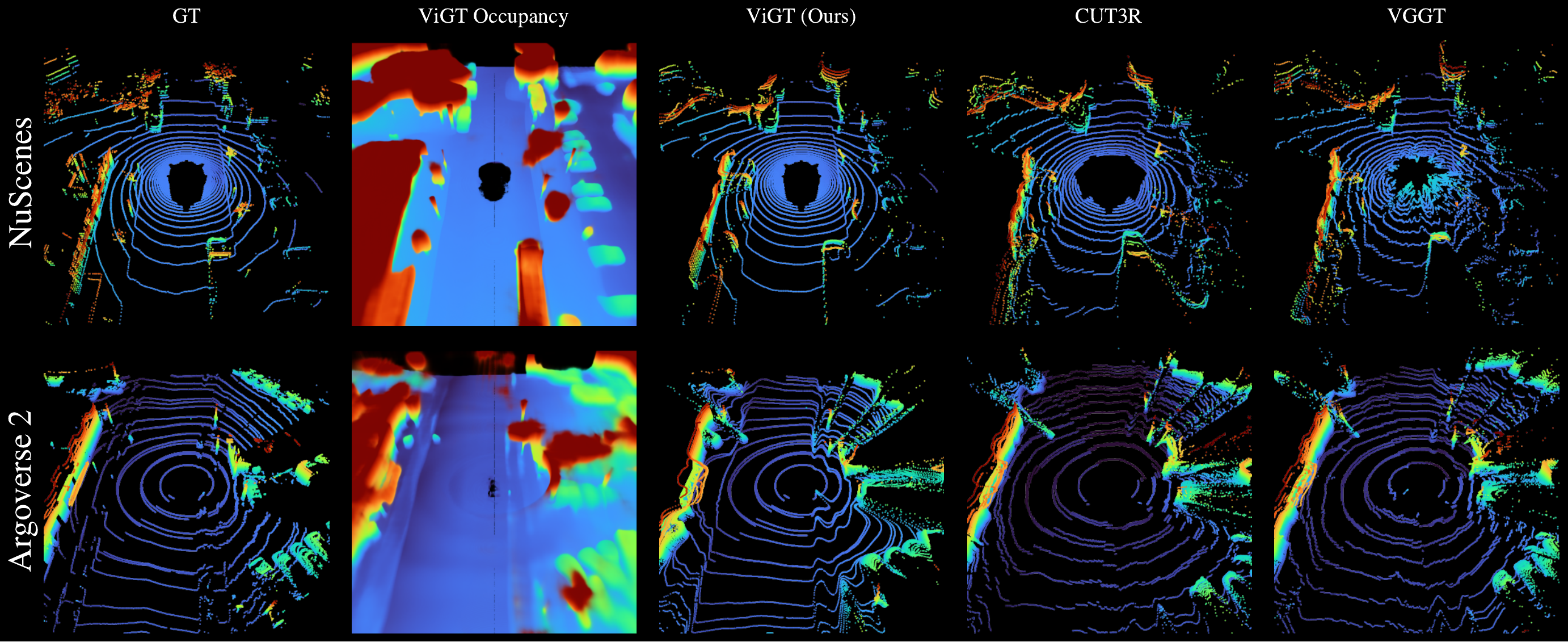}
    \caption{Visual comparison of GT and predicted point clouds across the top-3 methods ranked by Average Rank (Ours, Cut3R, VGGT). Our method produces more geometrically accurate point clouds. We also visualize a render of the predicted occupancy, demonstrating accurate capture of continious geometric structures.. The color encodes height along the z-axis.}
    \label{fig:compare_pcs}
\end{figure*}
\paragraph{Baselines}
Table~\ref{tab:results} presents a comprehensive comparison of our method against state-of-the-art approaches. The evaluated methods can be categorized into two groups: (1) general geometry methods that produce image-aligned representations, and (2) self-driving domain methods that estimate 3D scene representation. For general geometry methods (VGGT~\cite{vggt}, DUSt3R~\cite{dust3r2023}, Mast3R~\cite{mast3r}, Monst3R~\cite{monst3r}, Stream3R~\cite{stream3r}, Cut3R~\cite{cut3r}, and DA3~\cite{depthanythingv32025}), we evaluate them using two variants: depth maps and pointmaps, both filtered by LiDAR ray masks to extract valid 3D points for comparison. Note that we use metric-scale DA3 that only produces depth maps, so we evaluate it using the depth map variant only. For scene-centric methods including ours and RenderOcc~\cite{pan2024renderocc}, we render depth by LiDAR rays using Eq.~(\ref{eq:ray_rendering}).
\paragraph{Experimental results.}
As shown in Table~\ref{tab:results}, our method achieves the best average rank (1.8) across all metrics and datasets. Figure~\ref{fig:compare_pcs} provides a visual comparison of the top-3 methods by Average Rank. We outperform all competing methods in both AbsRel and CD metrics on NuScenes and Argoverse 2. On NuScenes, we achieve AbsRel of 0.068 and CD of 1.807, representing improvements of 0.105 in AbsRel and 1.83 in CD over the second-best method (Stream3R with depth-maps). On Argoverse 2, we achieve AbsRel of 0.131, outperforming the best competitor by 0.031 (Stream3R with depth-maps); and we achieve CD of 2.965, outperforming the best competitor by 0.716 (Cut3R with depth-maps). On Waymo, we achieve the second-best CD performance (2.431), trailing the best method by only 0.05. On ONCE and NuPlan, we achieve the best AbsRel performance, with third-best and second-best CD performance respectively.

Compared to RenderOcc, the most similar autonomous driving model, we consistently outperform it across all evaluated metrics. These results demonstrate the effectiveness of our calibration-free, self-supervised approach in accurately estimating 3D geometry from camera rig images across diverse autonomous driving datasets.

\begin{table}[!t]
    \centering
    \small
    \resizebox{0.45\textwidth}{!}{%
    \begin{tabular}{lccc}
        \toprule
        Component & \multicolumn{1}{c}{Point Est.} & \multicolumn{2}{c}{Voxel Occ.} \\
        \midrule
        \textit{Implicit BEV Projector Architecture} & CD$\downarrow$ & F1-score$\uparrow$ & IoU$\uparrow$ \\
        \quad CA & 3.051 & 0.697 & 0.547 \\
        \quad CA + CA & \textbf{2.699} & \textbf{0.713} & \textbf{0.566} \\
        \quad CA + SA + CA & 2.814 & 0.697 & 0.548 \\
        \midrule
        \textit{Encoder Output Layers} & CD$\downarrow$ & F1-score$\uparrow$ & IoU$\uparrow$ \\
        \quad Last 4 layers & \textbf{2.699} & \textbf{0.713} & \textbf{0.566} \\
        \quad Layers 5, 11, 16, 23 & 3.093 & 0.70 & 0.55 \\
        \midrule
        \textit{Query Sampling Strategy} & CD$\downarrow$ & F1-score$\uparrow$ & IoU$\uparrow$ \\
        \quad Random & 2.895 & 0.701 & 0.552 \\
        \quad Stratified & 3.039 & 0.697 & 0.547 \\
        \quad Stratified + Sym. Interval & \textbf{2.699} & \textbf{0.713} & \textbf{0.566} \\
        \bottomrule
    \end{tabular}%
    }
    \caption{Ablation study of key design choices: BEV projector architecture, encoder layer selection, and query sampling strategy.}
    \label{tab:ablation}
\end{table}

\subsection{Ablation Studies}
We evaluate three main components: the implicit BEV projector architecture, encoder output layer selection, and query sampling strategy (see Table~\ref{tab:ablation}.). All ablation experiments are trained on NuScenes only, using 4 GPUs for 100k iterations.

\paragraph{BEV Projector Architecture}
We compare three variants of implicit BEV projection module: a single cross-attention block (CA), two sequential cross-attention blocks (CA + CA), and a variant with self-attention (CA + SA + CA). The CA + CA configuration achieves the best performance, demonstrating that two sequential cross-attention operations provide sufficient capacity to learn the geometric transformation without the added complexity of self-attention.

\paragraph{Encoder Layer Selection}
We evaluate which ViT-L encoder layers to project to BEV space for DPT aggregation and upsampling. We compare using the last four layers versus default ViT-L output layers (5, 11, 16, 23). The last four layers selection shows better performance.

\paragraph{Query Sampling Strategy}
We ablate the strategy for sampling query points along LiDAR rays for occupancy supervision. We compare random sampling, stratified bin-based sampling, and stratified sampling augmented with symmetric interval sampling near object boundaries. The stratified + symmetric interval strategy performs best, confirming that explicit supervision near object boundaries is crucial for accurate occupancy estimation.

\section{Conclusion}

In this paper, we propose ViGT, a scalable vision-based geometric model for autonomous driving that learns a continuous 3D occupancy field directly from uncalibrated multi-camera images. ViGT is trained jointly on five large-scale datasets using a self-supervised learning procedure based on synchronized image–LiDAR pairs. We show that the resulting unified representation can be directly rendered for multiple downstream tasks, including pointmap estimation and occupancy prediction, without task-specific retraining. Extensive evaluations demonstrate that ViGT generalizes effectively, achieving state-of-the-art performance on pointmap estimation across five datasets with diverse camera rig configurations while remaining competitive with supervised methods on voxel-based occupancy benchmark. 
We consider such models representing a promising research direction toward building foundational geometric perception systems for autonomous driving applications.

\bibliographystyle{named}
\bibliography{ijcai26}

\onecolumn
\begin{appendices}
\begin{figure*}[!b]
    \centering
    \includegraphics[width=\linewidth,height=0.32\textheight,keepaspectratio]{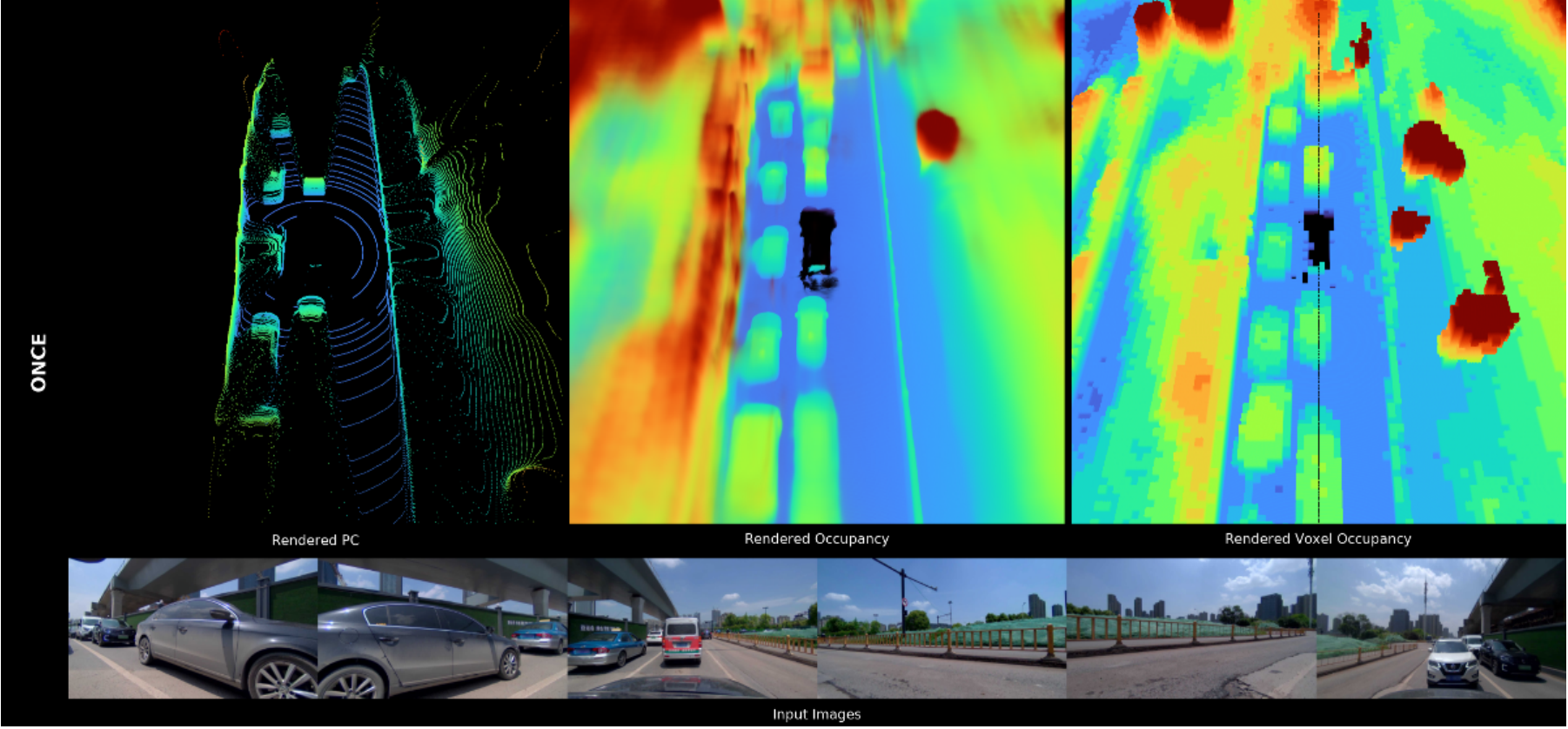}
    \caption{Our model estimates a continuous occupancy field directly from raw camera images, enabling flexible transformation into three downstream representations: rendered point clouds, rendered occupancy fields, and rendered voxel occupancy grids. This figure shows the input images (bottom row) and the corresponding three geometric representations derived from them.  Results are shown on the Once dataset. (Part 1 of 3)}
    \label{fig:mv0}
\end{figure*}

\section{Model Details}
Our model operates within a scene ROI box defined as $[-40, 40]$ m $\times$ $[-40, 40]$ m $\times$ $[-1, 5.4]$ m in the ego coordinate frame.

\textbf{Implicit BEV Projection.} The calibration-free camera-to-BEV projection module employs two sequential cross-attention blocks for each encoder layer. Each cross-attention block operates between image patch tokens and a set of BEV latent queries. The BEV query space is defined as a grid of resolution $32 \times 32$, resulting in 1024 BEV queries per layer. Each BEV query has a dimension of 1024. The four layer-specific BEV representations are concatenated along the feature dimension and upsampled using DPT~\cite{dpt2021} to produce the final BEV representation at resolution $256 \times 256$ with 256-dimensional features.

\textbf{Training Hyperparameters.} We use balanced query sampling along LiDAR rays, dividing the free space segment $[0, d)$ into $K = 5$ bins for stratified sampling. The positive sampling thickness hyperparameter $\tau$ is set to 0.1 meters. During training, we sample 150K positive and 150K negative query points per sample, where negative queries consist of 120K points from stratified bins and 30K points from the symmetric interval $[d-\tau, d)$ near object surfaces.

\section{Evaluation Details}

\subsection{Point Map Estimation}
Point maps for our model are rendered from the implicit occupancy field by sampling points along each LiDAR ray with a step size of 0.05 m. For baseline comparisons, point maps and depth maps from baseline methods were filtered by computing a visibility mask using LiDAR rays projected into camera views via camera calibrations.

\subsection{Occupancy Estimation}
For occupancy voxel grid evaluation, we aggregate point-wise occupancy predictions of our model from 8 points sampled uniformly within each voxel by taking the maximum occupancy probability.

\section{Additional Qualitative Results}

\subsection{Multi-Representation Visualization}
\begin{figure*}[!ht]
    \centering
    \includegraphics[width=\linewidth,height=0.32\textheight]{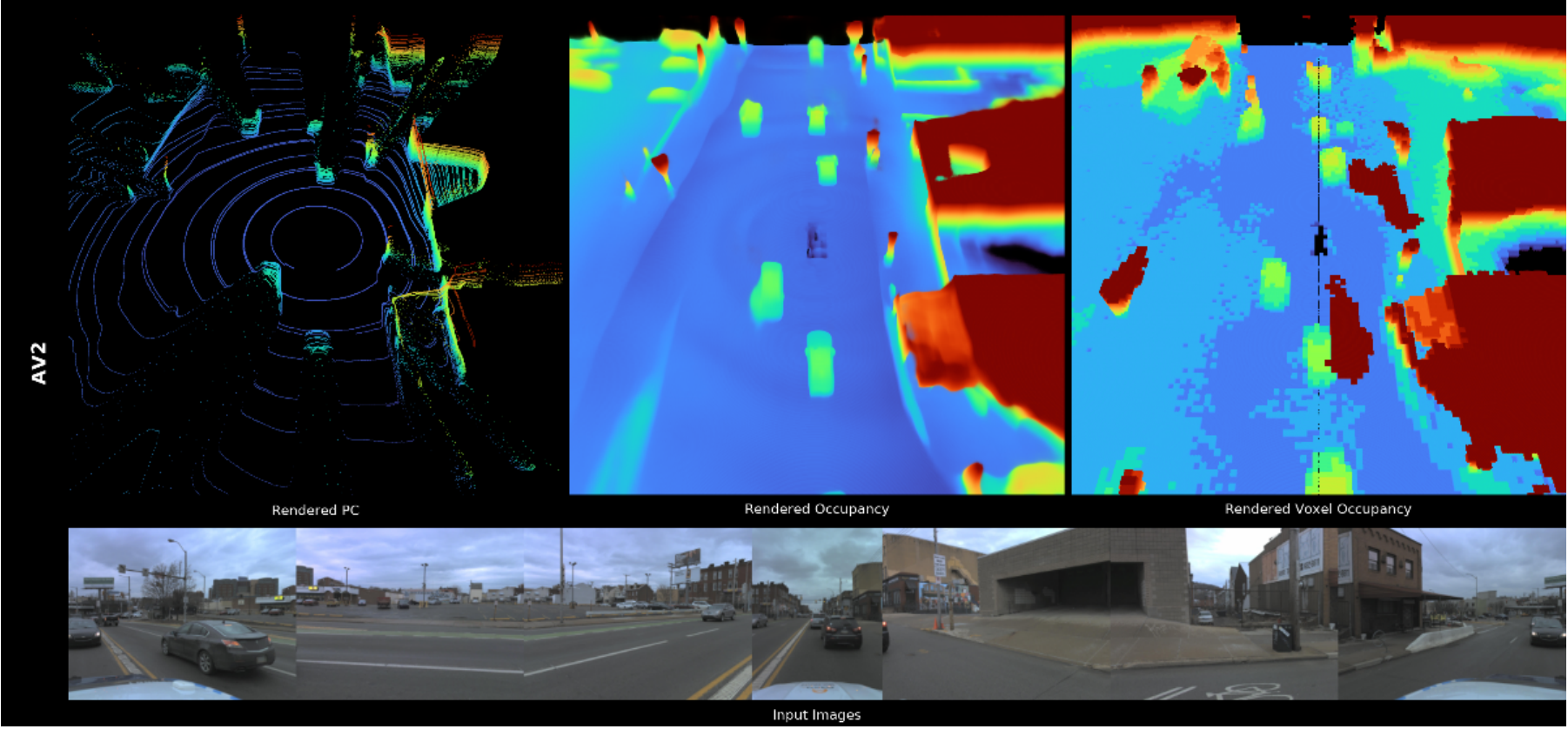}
    \includegraphics[width=\linewidth,height=0.32\textheight]{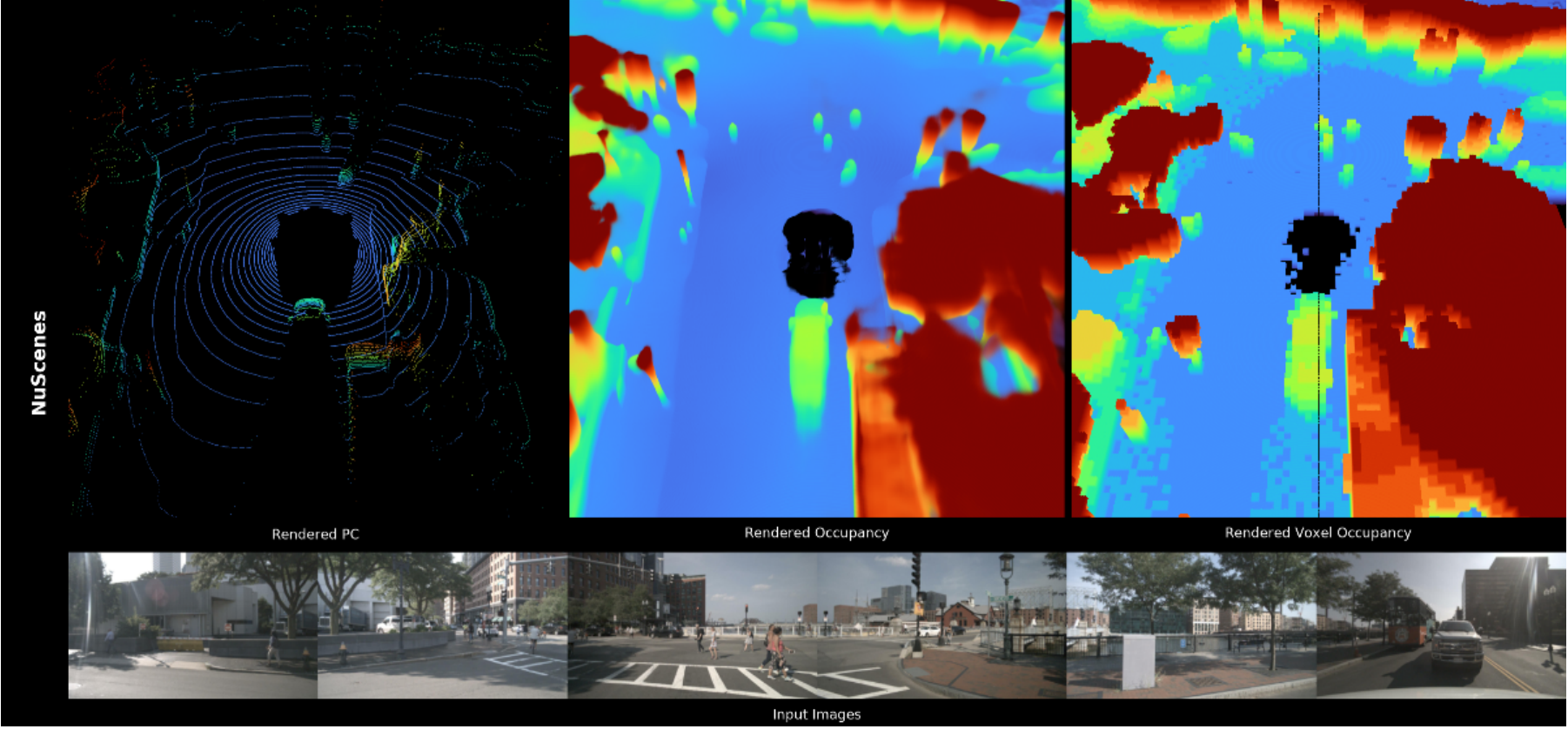}
    \caption{Our model estimates a continuous occupancy field directly from raw camera images, enabling flexible transformation into three downstream representations: rendered point clouds, rendered occupancy fields, and rendered voxel occupancy grids. This figure shows the input images (bottom row) and the corresponding three geometric representations derived from them. Results are shown on AV2 and nuScenes datasets. (Part 2 of 3)}
    \label{fig:mv1}
\end{figure*}

\begin{figure*}[!ht]
    \centering
    \includegraphics[width=\linewidth]{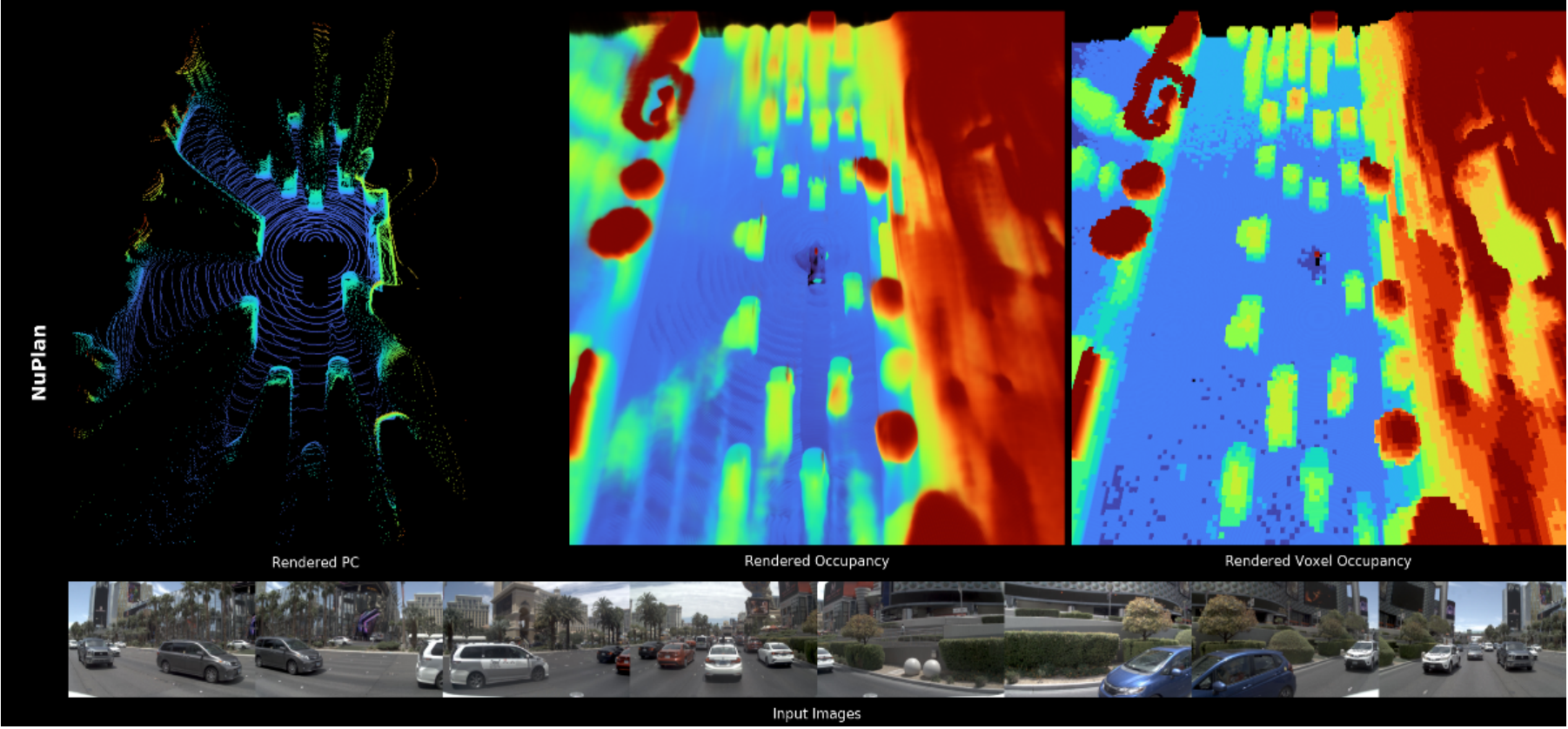}
    \includegraphics[width=\linewidth,height=0.42\textheight,keepaspectratio]{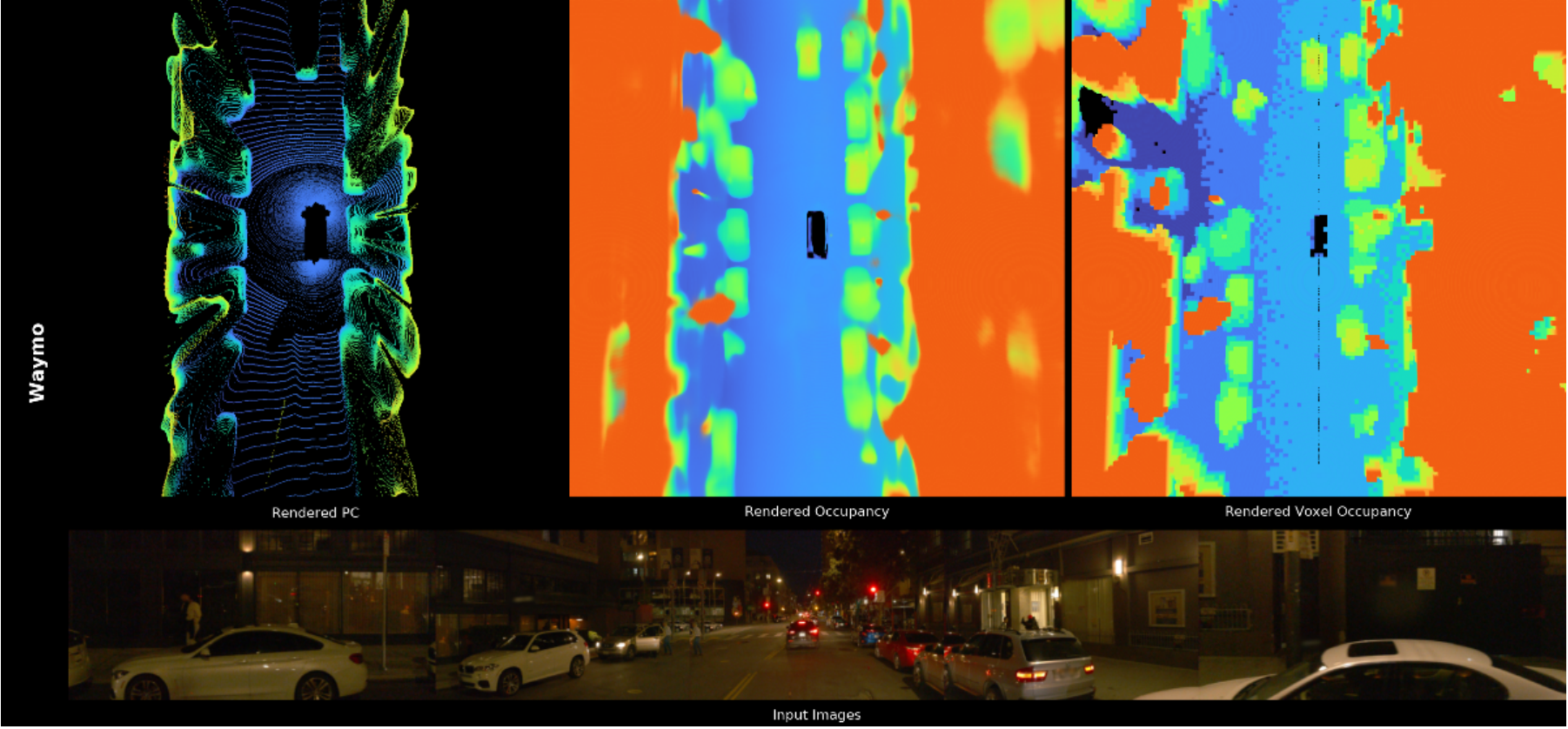}
    \caption{Our model estimates a continuous occupancy field directly from raw camera images, enabling flexible transformation into three downstream representations: rendered point clouds, rendered occupancy fields, and rendered voxel occupancy grids. This figure shows the input images (bottom row) and the corresponding three geometric representations derived from them.  Results are shown on NuPlan and Waymo datasets. (Part 3 of 3)}
    \label{fig:mv2}
\end{figure*}

Our model estimates a continuous occupancy field directly from camera images, enabling flexible transformation into three
downstream representations: rendered point clouds, rendered occupancy fields, and rendered voxel occupancy grids. 
Figures~\ref{fig:mv0},\ref{fig:mv1} and~\ref{fig:mv2} illustrate this flexibility by showing rendered point clouds, rendered occupancy fields, and rendered voxel grids across five datasets (AV2, nuScenes, ONCE, NuPlan, and Waymo).

\subsection{Robustness to Camera Dropout}
\begin{figure*}[b]
    \centering
    \includegraphics[width=\linewidth]{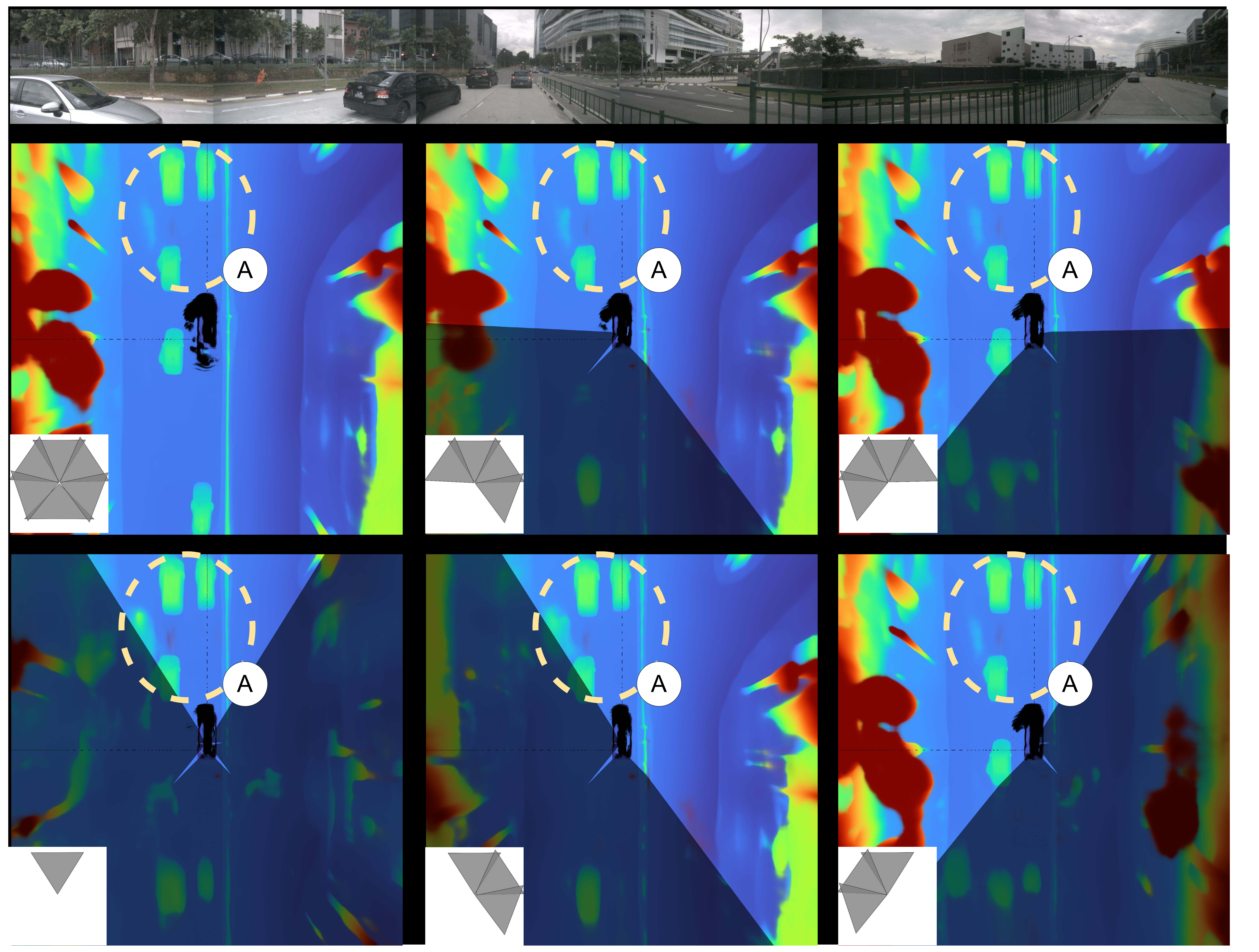}
    \caption{
    Model occupancy predictions for different subsets of surround-view images (top row). 
    In each illustration, input images are indicated by corresponding cameras frustums in the lower-left corner, while invisible regions are tinted. 
    The model provides consistent predictions within regions of visibility (A,B,C) with some or even most of the cameras removed.
    Part 1 shows sample scene from NuScenes.
    (Part 1 of 2)}
    \label{fig:drop_nusc}
\end{figure*}

\begin{figure*}[]
    \centering
    \includegraphics[width=\linewidth]{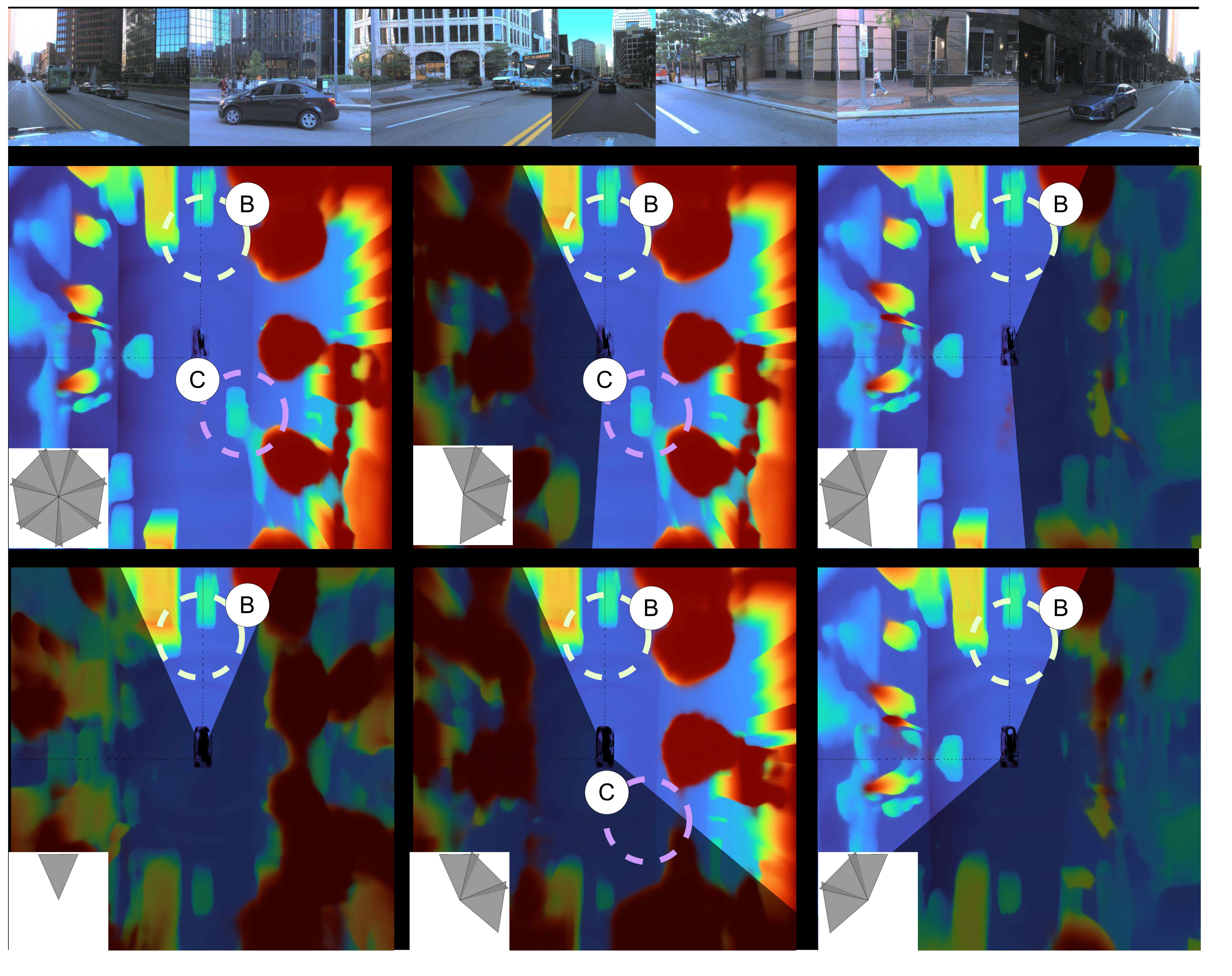}
    \caption{
    Model occupancy predictions for different subsets of surround-view images (top row). 
    In each illustration, input images are indicated by corresponding cameras frustums in the lower-left corner, while invisible regions are tinted. 
    The model provides consistent predictions within regions of visibility (A,B,C) with some or even most of the cameras removed.
    Part 2 shows sample scene from ArgoVerse2.
    (Part 2 of 2)}
    \label{fig:drop_av2}
\end{figure*}

To demonstrate the robustness of our calibration-free approach, we evaluate the model's performance when cameras are dropped from the input (Figure~\ref{fig:drop_nusc}, Figure~\ref{fig:drop_av2}).

\subsection{Visual comparison on pointmap estimation}
\begin{figure*}[]
    \centering
    \includegraphics[width=\linewidth]{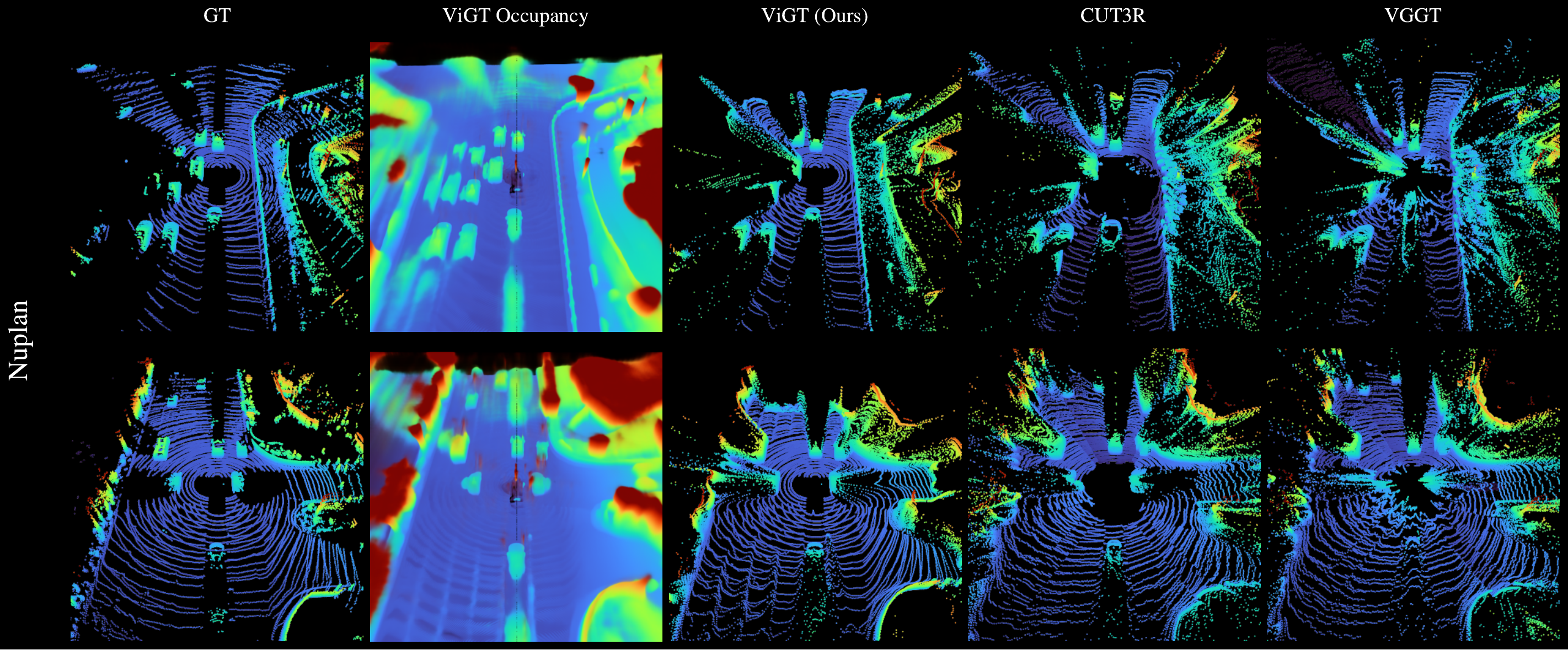}
    \caption{Visual comparison of GT and predicted point clouds across the top-3 methods ranked by Average Rank (Ours, Cut3R, VGGT) (continued). Our method produces more geometrically accurate point clouds. We also visualize a render of the predicted occupancy, demonstrating accurate capture of continious geometric structures. The color encodes height along the z-axis. (Part 2 of 2)}
    \label{fig:compare_pcs_continued}
\end{figure*}

We compare our method against the top-ranked competing methods from Table~\ref{tab:results} (Cut3R, VGGT) across AV2, nuScenes, ONCE, and NuPlan datasets (Figures~\ref{fig:compare_pcs},\ref{fig:compare_pcs_continued}), demonstrating superior geometric accuracy in point cloud predictions.


\begin{figure*}[]
    \centering
    \includegraphics[width=\linewidth]{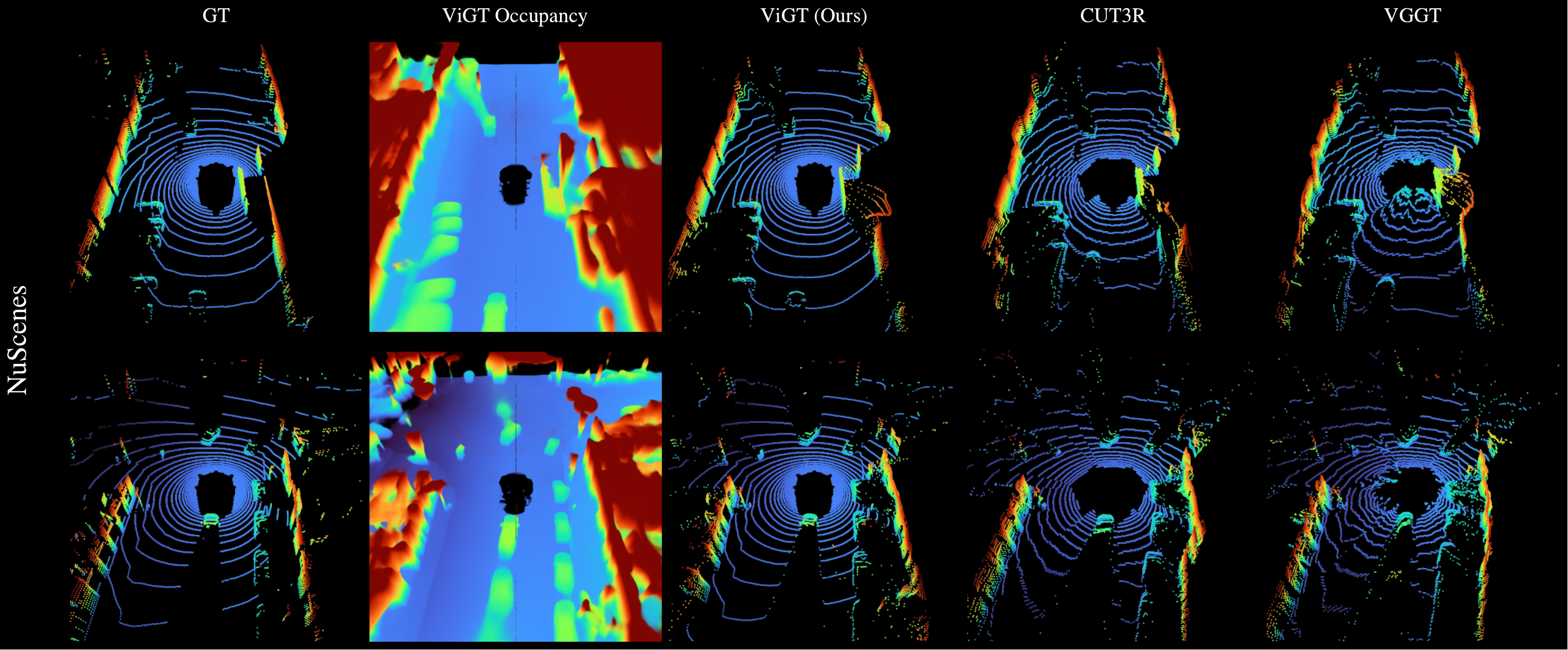}
    \includegraphics[width=\linewidth]{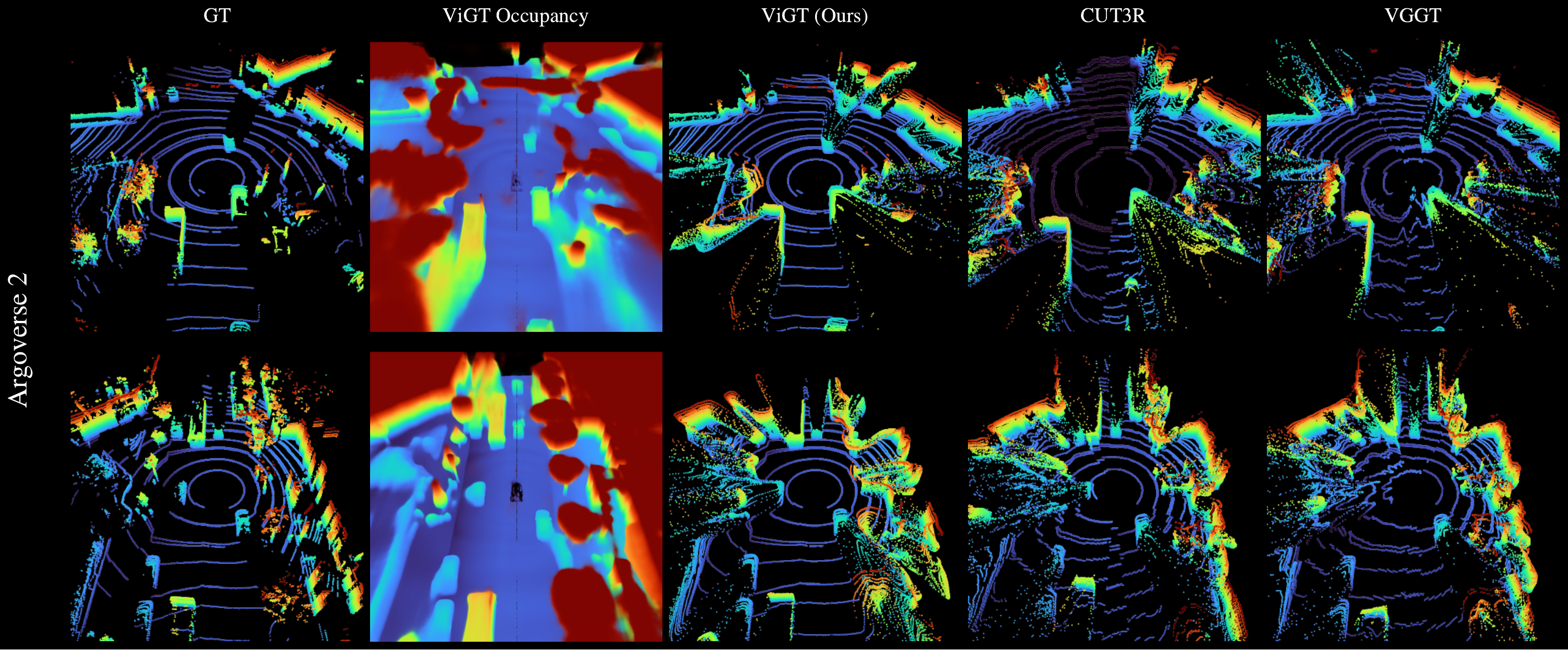}
    \includegraphics[width=\linewidth]{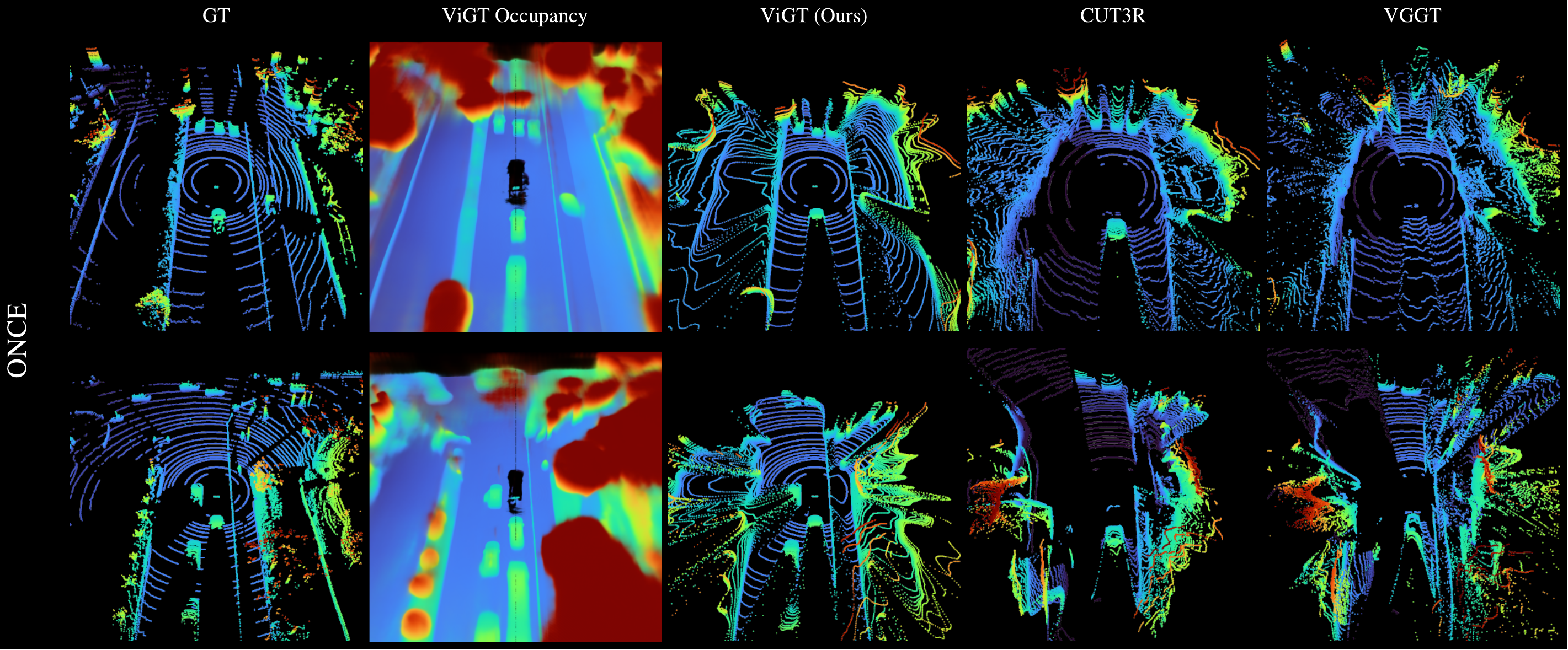}
    \caption{Visual comparison of GT and predicted point clouds across the top-3 methods ranked by Average Rank (Ours, Cut3R, VGGT). Our method produces more geometrically accurate point clouds. We also visualize a render of the predicted occupancy, demonstrating accurate capture of continious geometric structures. The color encodes height along the z-axis. (Part 1 of 2)}
    \label{fig:compare_pcs}
\end{figure*}

\end{appendices}

\end{document}